\documentclass[runningheads]{llncs}

\usepackage{eccv}

\usepackage{eccvabbrv}

\usepackage{graphicx}
\usepackage{booktabs}

\usepackage[accsupp]{axessibility}  

\usepackage{hyperref}

\usepackage{orcidlink}

\definecolor{mygray}{rgb}{0.95,0.95,0.95}
\definecolor{mypink}{rgb}{1,0.49,0.51}
\definecolor{myorange}{rgb}{1,0.75,0.40}
\usepackage{xcolor}
\usepackage{color, colortbl}
\usepackage{soul} 
\definecolor{mylightorange}{RGB}{255,242,209}
\definecolor{mylightpurple}{RGB}{224,227,253}
\definecolor{mygreen}{RGB}{170,227,150}
\definecolor{mylightblue}{RGB}{126,164,193}
\DeclareMathOperator*{\argmax}{arg\,max}
\newcommand{\innerproduct}[2]{\langle #1, #2 \rangle}
\begin{document}

\title{Data Augmentation via Latent Diffusion for Saliency Prediction} 
\author{Bahar Aydemir\inst{1}\orcidlink{0000-0001-5202-5240} \and
Deblina Bhattacharjee\inst{1}\orcidlink{0000-0002-0534-852X} \and
Tong Zhang\inst{1}\orcidlink{0000-0001-5818-4285} \and
Mathieu Salzmann\inst{1}\orcidlink{0000-0002-8347-8637} \and
Sabine Süsstrunk\inst{1}\orcidlink{0000-0002-0441-6068}}
\authorrunning{B.~Aydemir et al.}
\institute{School of Computer and Communication Sciences, EPFL, Switzerland \\ \email{\{name\}.\{surname\}@epfl.ch}}
\maketitle
  \begin{figure*}[h!]
\centering
\includegraphics[width=0.92\columnwidth]{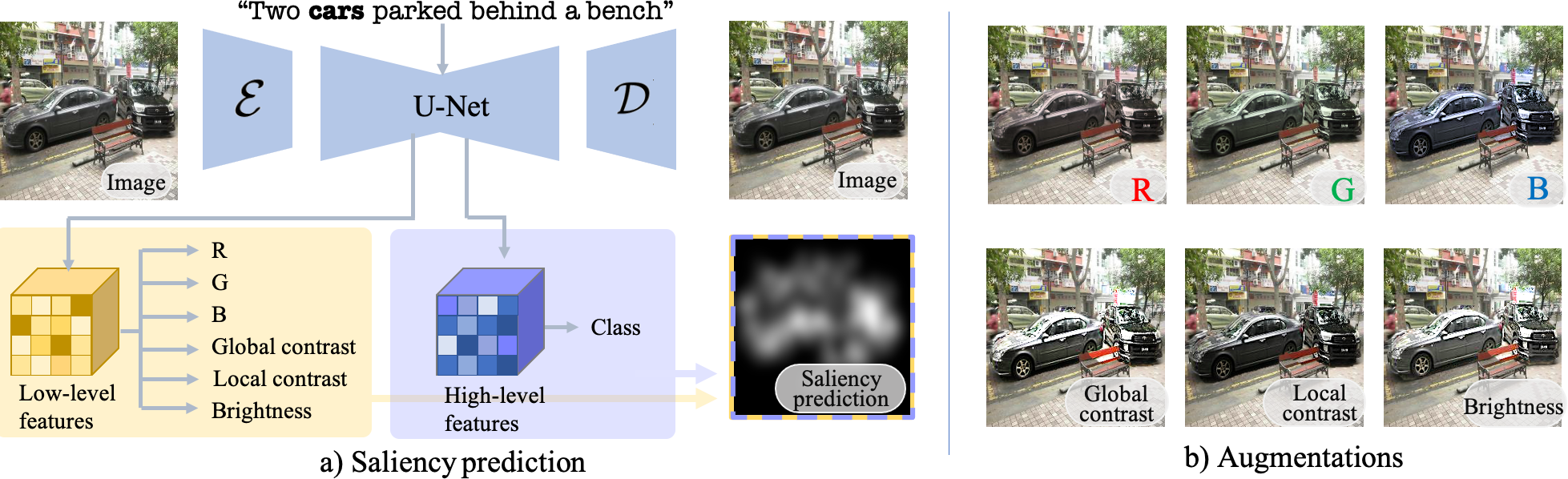}
    \caption{\textbf{Overview of our data augmentation and saliency prediction method.} We use photometric and semantic properties such as color, contrast brightness, and class to construct \sethlcolor{mylightorange}\hl{low-level} and \sethlcolor{mylightpurple}\hl{high-level} features from the intermediate U-Net features. We leverage these features to estimate saliency and generate image edits to augment training data for saliency prediction. We use a saliency-guided cross-attention mechanism between the input image and the text prompt to perform localized edits. }
    
    \label{fig:teaser}
\end{figure*}
\begin{abstract}
Saliency prediction models are constrained by the limited diversity and quantity of labeled data. Standard data augmentation techniques such as rotating and cropping alter scene composition, affecting saliency. We propose a novel data augmentation method for deep saliency prediction that edits natural images while preserving the complexity and variability of real-world scenes. Since saliency depends on high-level and low-level features, our approach involves learning both by incorporating photometric and semantic attributes such as color, contrast, brightness, and class. To that end, we introduce a saliency-guided cross-attention mechanism that enables targeted edits on the photometric properties, thereby enhancing saliency within specific image regions. Experimental results show that our data augmentation method consistently improves the performance of various saliency models. Moreover, leveraging the augmentation features for saliency prediction yields superior performance on publicly available saliency benchmarks. Our predictions align closely with human visual attention patterns in the edited images, as validated by a user study. Our code is publicly available on \href{https://github.com/IVRL/Augsal}{GitHub}\footnotemark.
. 
\end{abstract}

\section{Introduction}
\par Visual saliency prediction, with applications in image and video compression \cite{saliencycompression}, and image enhancement \cite{image_enhance1, image_enhance2}, aims to identify the regions within an image that attract the human gaze. 
The taxonomy of saliency estimation covers both bottom-up and deep-learning approaches. Bottom-up methods use hand-crafted features to estimate the saliency of a region, whereas the deep learning approaches aim to benefit from large-scale data.
Recent methods incorporate various priors such as object dissimilarity~\cite{objdissim}, temporal information~\cite{aydemir2023tempsal}, and inter-object relationships~\cite{external-knowledge-saliency} alongside large-scale data to enhance performance. However, the limited size of available datasets poses a significant challenge to these learning-based methods. The most extensive saliency dataset contains only 10k training images~\cite{salicon}, in contrast to the millions of images in classic computer vision datasets \cite{deng2009imagenet,Sun2017RevisitingUE}. The reason is that collecting ground-truth saliency data through psychophysical experiments is both costly and time-consuming. Hence, collecting saliency annotations on the scale of millions is not practically feasible. In the general computer vision context, limited access to data is typically overcome via data augmentation. However, standard data augmentation techniques such as cropping, rotation, and shearing, used in tasks like image classification and segmentation, are not saliency invariant \cite{Borjigaze}; such manipulations also alter human gaze patterns. This makes the existing data augmentation techniques ineffective, as saliency is not invariant under such transformations.
\footnotetext{https://github.com/IVRL/Augsal}
This raises a crucial question: Is it possible to manipulate image saliency in a \textit{predictable} manner, such as \textit{enhancing the saliency of a region} while \textit{maintaining the integrity of the rest of the scene}? That is, can we change the saliency of a region in an expected direction? To answer this question, we refer to the factors influencing saliency. Studies in cognitive science \cite{Wolfe2004} have demonstrated that saliency is affected by both low-level features, such as color~\cite{color2BAUER19961439,color}, contrast\cite{contrastPashler2004}, and brightness~\cite{brightness}, and high-level features, such as semantics~\cite{objectsHelpGazePrediction,objectness}. In essence, the saliency of a region is scene-dependent and can vary dramatically from one image to another.  
Although synthetic datasets \cite{Berga2019SID4VAMAB} featuring basic shapes in various colors, sizes, and orientations study the impact of the above factors on saliency, they fall short in complex everyday scenes with real objects. Regarding the question posed earlier, \textit{yes, we can change saliency predictably}. In this work, we build on this idea by selecting a region and increasing its saliency. The edited region's saliency should be higher than the original. Thus, this transformation, with a known direction of saliency change, is suitable for data augmentation in saliency prediction.

In our work, we address these limitations by introducing an image editing strategy that allows us to alter a single factor in a scene while keeping the others constant. This allows us to control (enhance or decrease) and interpret the saliency of regions within an image while preserving the integrity of the rest of the real-world visual scene. 
To achieve this, we use readout layers to constrain the image edits within a range for the edited properties while increasing the saliency over the selected region. This approach enables us to alter a single aspect of the scene while modifying the saliency in a desired direction. This process of editing the image and subsequently, modifying the saliency allows our method to generate image-saliency annotation pairs, thereby addressing the lack of large-scale data for learning-based saliency approaches.
 Moreover, we employ a vision-language-based cross-attention mechanism~\cite{Tan2019LXMERTLC} that localizes the target image regions for editing, thereby eliminating the need for user-provided input masks as required in~\cite{visualdist,cvpr2023sal}. Thus, the cross-attention mechanism completely automates our editing method for saliency prediction, making it suitable for data augmentation. 

To edit the image in a controllable and interpretable manner for visual saliency, we employ the widely-used diffusion model~\cite{dhariwal2021diffusion}. Recent text-to-image generation models have proven their ability to generate diverse and creative images. They enable image generation by conditioning on the input text prompts. They have also been adapted for dense prediction tasks such as depth estimation~\cite{saxena2023monocular,duan2023diffusiondepth,ji2023ddp} and segmentation~\cite{tan2023diffss,amit2022segdiff}, showcasing their transferability. However, saliency prediction presents a unique set of challenges, as it is highly dependent on both the semantic content and the low-level details within the images, requiring an understanding of visual importance that goes beyond structural coherence. Since diffusion models~\cite{rombach2021ldm} are trained on large and diverse data, they are suitable to exploit their pre-trained knowledge in saliency prediction.

We summarize our contributions as follows:
\begin{itemize}
\item  We introduce a data augmentation method that enriches the training data while predictably modifying image saliency. 
\item Our approach introduces multi-level feature readouts, allowing us to exploit knowledge from the Stable Diffusion~\cite{rombach2021ldm} architecture without retraining for image saliency prediction. 
\item We introduce controllability in saliency by editing photometric properties, namely, contrast, brightness, and color, in a shared latent feature space of generated image edits and saliency using cross-attention mechanisms. 
\item Our model is interpretable in the type of image edits that lead to enhanced saliency, which, in turn, aligns better with human visual attention, as we show via a user study.
\end{itemize}
Our experiments evidence that our image editing-based data augmentation strategy consistently improves the state-of-the-art saliency predictors. Furthermore, our novel diffusion-based saliency estimation approach outperforms the existing saliency models. 
\section{Related Work}

\subsection{Saliency Prediction Models}
Convolutional neural networks (CNNs) have attained significant popularity in deep saliency prediction, as evidenced by numerous studies~\cite{kummerer2015deep,eml,aydemir2023tempsal,simplenet,deepgaze2} and pioneered by~\cite{vigsal}.  Notably, Kümmerer et al.~\cite{kummerer2015deep} have shown that CNNs trained for object classification greatly enhance saliency prediction, aligning with Judd et al.'s~\cite{Judd_2009} earlier findings using bottom-up detectors. This approach has been adopted by leading deep saliency prediction networks like~\cite{saliconmodel,deepgaze2,kruthiventi2015deepfix,mlnet,sam}
and further developed by EML-Net \cite{eml}, which integrates features from multiple CNN backbones for object classification. Linardos et al.~\cite{linardos} also explored various object classification backbones for saliency prediction. Recently,~\cite{TranSalNet} has incorporated transformer blocks to encode long-range relationships, although still relying on a CNN encoder. All these methods transfer the learned encodings from \textit{object recognition} (based on ImageNet~\cite{deng2009imagenet}) to saliency prediction.
Differently, in this work, we leverage the encodings from a large \textit{vision-language-based} model to predict saliency. In particular, we use a pre-trained and frozen Stable Diffusion~\cite{rombach2021ldm} model that encodes information from 2.3 billion images. 
We create multi-level features from this pre-trained model for generating image edits and predicting saliency. The use of diffusion models allows our approach to be both controllable and interpretable.  

\subsection{Diffusion Models on Dense Prediction Tasks }
 Diffusion-based approaches gradually corrupt an image with different levels of noise to train denoiser models. Consequently, the model learns how to denoise a random noise into a high-quality synthetic image~\cite{ermon2019}. 
 Diffusion models have demonstrated remarkable results in multiple fields including audio processing, natural language processing, and video generation \cite{schneider2023archisound,zhu2023diffusion,ho2022video}, as well as in multiple tasks such as image generation \cite{chefer2023attend,zhang2023controlnet,rombach2021ldm}, semantic segmentation~\cite{tan2023diffss,amit2022segdiff}, and depth estimation~\cite{saxena2023monocular,duan2023diffusiondepth,ji2023ddp}, among others.

 All of these methods achieve state-of-the-art performance on their respective benchmarks by exploiting structural coherence in the scene which is encoded within the diffusion model. However, saliency prediction requires understanding the scene beyond structural coherence.
 Therefore, we exploit the pre-trained diffusion model by generating image edits that control the saliency of a desired target region. Guided by these image edits, our method predicts saliency in alignment with human visual attention.
To the best of our knowledge, this is the first work to predict and control saliency in a diffusion framework.

\subsection{Cross-attention to Generate Image Edits}
Generative models are well-known for their capability to edit images. Following the seminal work on cross-attention~\cite{Tan2019LXMERTLC} that exploits language and vision encodings, many works have employed the cross-attention mechanism within the diffusion paradigm to generate image edits~\cite{chefer2023attend, brooks2023instructpix2pix, hertz2022prompt, rombach2021ldm}.
In particular, \cite{chefer2023attend} introduces a semantically guided cross-attention mechanism to generate images. Differently, Hertz et al.~\cite{hertz2022prompt} presents a text-driven image editing method using cross-attention. Subsequently, Brooks et al.~\cite{brooks2023instructpix2pix} learn to follow image editing instructions given an input image and a text prompt. While these works reason about the cross-attention between text and image, they do not consider photometric properties such as contrast or salient regions to generate image edits. Since contrast and saliency are significant cues to generate image edits, we introduce a \textit{saliency-guided cross-attention} mechanism within our diffusion paradigm. Using the cross-attention mechanism makes our editing method for saliency prediction completely automated and thus well-suited for data augmentation.

 \begin{figure*}[t]
  \centering
  \includegraphics[width=0.9\textwidth]{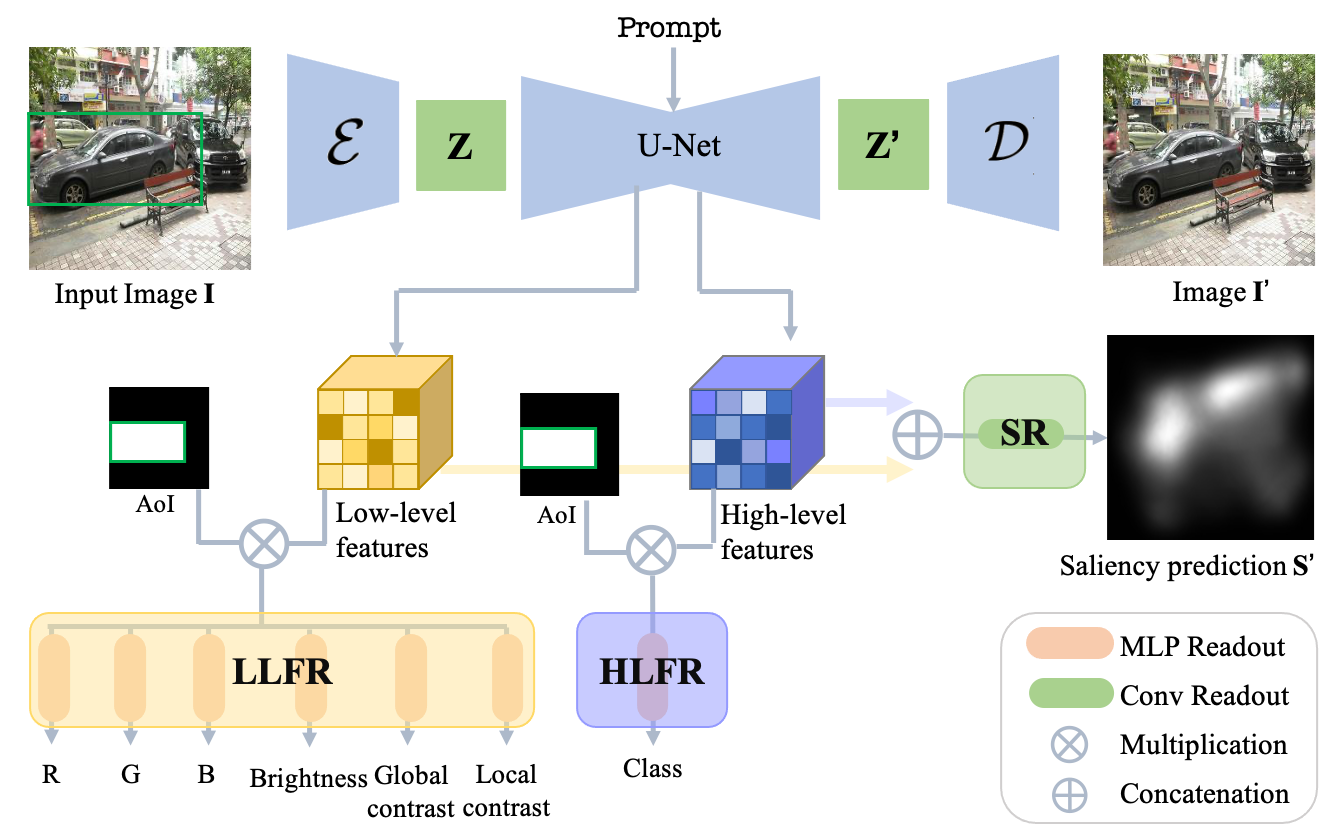}
  \caption{\small{\textbf{Overview of the proposed architecture in training.} We use an \textbf{\textcolor{mylightblue}{encoder}} (\(\mathcal{E}\)), a \textbf{\textcolor{mylightblue}{decoder}} (\(\mathcal{D}\)), and a denoising \textbf{\textcolor{mylightblue}{U-Net}} from Stable Diffusion ~\cite{rombach2021ldm}. We invert the input image to extract encoded representations from the U-Net to construct \sethlcolor{mylightorange}\hl{low-level} and \sethlcolor{mylightpurple}\hl{high-level} features. We train the Low-Level Feature Readout (LLFR) and High-Level Feature Readout (LLFR) modules using related photometric and semantic properties inside the area of interest (AoI), respectively. Finally, we concatenate the \sethlcolor{mylightorange}\hl{low-level} and \sethlcolor{mylightpurple}\hl{high-level} features to predict the saliency map (\(\mathbf{S}'\)) using the Saliency Readout (SR) module. \sethlcolor{mygreen}\hl{\textbf{Z}} and \sethlcolor{mygreen}\hl{\textbf{Z'}} represent the encoded and denoised image latent vectors respectively.}}
  \label{fig:train-figure}
\end{figure*}

\section{Methodology}

Our approach is depicted in Figure 2. We incorporate the encoder (\(\mathcal{E}\)), the decoder (\(\mathcal{D}\)), and the denoising U-Net from Stable Diffusion v1.5~\cite{rombach2021ldm}, all of which are frozen in our framework. Specifically, we start by encoding the input image and then extracting the intermediate representations from the middle layers of the denoising U-Net into two feature maps: high-level and low-level features. We use Low-Level Feature Readout(LLFR) and High-Level Feature Readout (HLFR) modules to learn desired photometric and semantic properties. 
The denoising U-Net denoises the latent feature \(\mathbf{Z}_t\) to produce \(\mathbf{Z}'\) and the image itself. Lastly, we concatenate the high and low-level features
and decode them with the Saliency Readout (SR) module to predict saliency. We detail the components of our model in what follows.
\subsection{Multi-level Features }
We start by encoding the input image and then extracting the intermediate representations from the middle layers of the denoising U-Net. We aggregate the intermediate representations with the bottleneck layers, as proposed in \cite{luo2023dhf}. The aggregated representations allow us to construct two feature maps comprising high-level and low-level features, respectively.
This multi-level representation allows our model to depend on both low-level and high-level features while predicting saliency.
The separation between those features allows them to optimize for the desired tasks without interacting with each other. We learn those features using the LLFR, HLFR, and SR modules as we explain in the next sections. 
\subsection{Low-Level Feature Readout (LLFR) Module} \label{sec:mlp-readouts}
We employ seven Multi-Level Perceptron based readout networks to access the intermediate representations of the desired photometric properties. Specifically, we utilize small readout networks that share the low-level features to interpret red, green, blue, brightness, and local and global contrast values. The choice of those properties is based on evidence showing that saliency is significantly influenced by color, contrast, and brightness~\cite{color,contrastPashler2004,brightness}.
To train those networks, we crop a random patch from the image and compute the related photometric properties inside this AoI as follows:\\
\textbf{R, G, B:} For each color channel, we calculate the average color in the patch as  
\[ \mu_{R} = \frac{1}{N} \sum_{i=1}^{N} \text{P}_{[0,i]}\;,\quad \mu_{G} = \frac{1}{N} \sum_{i=1}^{N} \text{P}_{[1,i]}\;,\quad  \mu_{B} = \frac{1}{N} \sum_{i=1}^{N} \text{P}_{[2,i]}\;, \] where $P_{[j,i]}$ denotes each pixel in the $j^{th}$ channel of the image patch.\\
\textbf{Brightness:} We first convert the image into grayscale and then compute the brightness as the mean intensity of the patch as
\[  \mu_{Br} = \frac{1}{N} \sum_{i=1}^{N} \text{P}_i\;, \]
where $P_{i}$ denotes each pixel in the image patch.\\
\textbf{Local Contrast:}  We compute the mean intensity (\(\mu\)) of the patch pixels, together with their variance (\(\sigma^2\)) and local contrast (c$_L$) as
    \[ \mu = \frac{1}{N} \sum_{i=1}^{N} \text{P}_i, \quad \quad  \sigma^2 = \frac{1}{N} \sum_{i=1}^{N} (\text{P}_i - \mu)^2, \quad \quad c_L = \sqrt{\sigma^2} \;. \]    
\textbf{Global Contrast:}    We compute the mean intensity (\(\mu\)) of the image pixels, as well as their variance (\(\sigma^2\)) and the global contrast (c$_G$) as
\[ \mu = \frac{1}{N} \sum_{i=1}^{N} \text{I}_i, \quad \quad \sigma^2 = \frac{1}{N} \sum_{i=1}^{N} (\text{P}_i - \mu)^2,\quad \quad c_G = \sqrt{\sigma^2}\;, \]\\
where $P_i$ and $I_i$ denote each pixel in the patch and image, respectively.
\subsection{ High-Level Feature Readout (HLFR) Module}
We use one convolution-based readout network to learn semantic information present in the scene. Specifically, we classify the object present in the area of interest. We aim to learn the semantics since saliency also depends on the semantics of the scene ~\cite{objectness,objectsHelpGazePrediction}.\\
\textbf{Classification:} 
We train the classification readout by using the object bounding boxes from the MS-COCO dataset \cite{mscoco}. We construct high-level features that contain the semantics of the objects in the selected region. Then, we pass these features to the saliency readout (SR) to provide information about the semantics present in the scene. 
\subsection{Saliency Readout (SR) Module }
The saliency readout network consists of 7 convolutional layers with SiLU~\cite{silu} activations. This module takes both the high-level and low-level features as input. We use skip connections to prevent the gradient of the features from vanishing. 
We decode the concatenated features to predict the final saliency map \(\mathbf{S}'\).
Equipped with high-level and low-level feature representations via readout networks, our approach can achieve both controllability and interpretability for saliency prediction.
\begin{figure*}[t]
  \centering
  \includegraphics[width=1\textwidth]{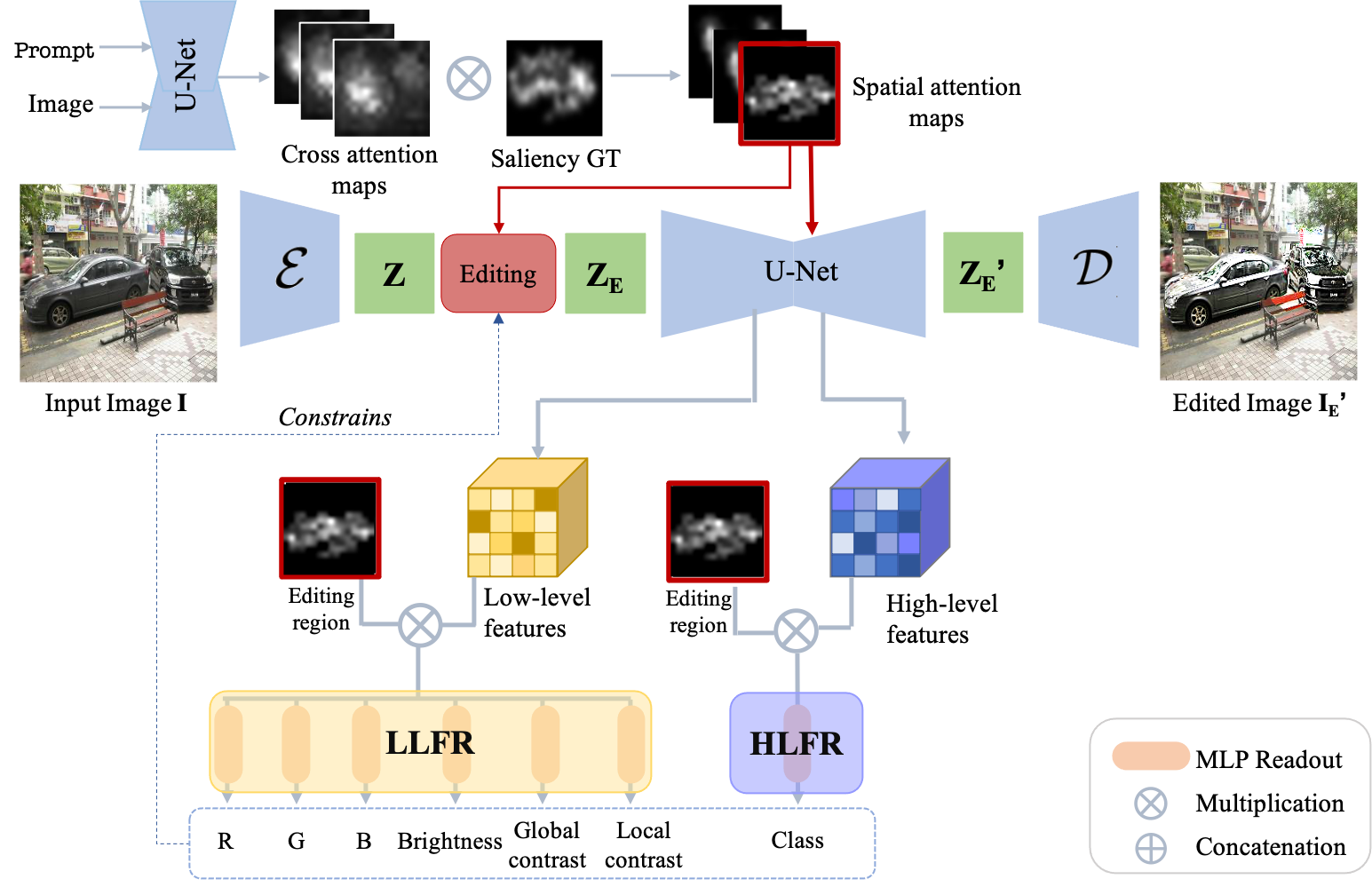}
  \caption{\textbf{Overview of the proposed image editing architecture.} We extract cross-attention maps between the input image and the prompt, and then multiply with the saliency map to create spatial attention maps. These maps highlight the intersections of salient regions with elements from the prompt. We inject this map into the denoising U-Net alongside the edited prompt. This integration modifies the latent features \( \mathbf{Z}' \) and the extracted multi-level features, resulting in the generating edits. We use frozen readout layers to constrain the edits in terms of those features. }
  \label{fig:model-editing}
\end{figure*}
\section{Image Editing for Data Augmentation}
Image editing requires two components: 1. Choosing the type of edits, and 2. locating the graphical elements on which the edits are carried out. In this work, we manually choose the type of edits and then apply them to the graphical elements by employing our saliency-guided cross-attention mechanism. We detail the different types of edits that we incorporate in our approach in Section \ref{sec:editing-types}. 
As shown in Figure \ref{fig:model-editing}, during the editing stage, we extract the cross-attention features between the input image and the text prompt. These features are multiplied with the ground-truth saliency maps from SALICON~\cite{salicon} to create the saliency-guided cross-attention maps. These maps reveal where the salient regions intersect with elements from the prompt. We select the spatial attention map with the highest sum, indicating the most salient area corresponding to a word in the prompt. Thus, we dub this process a saliency-guided cross-attention mechanism.
We leverage the obtained saliency-guided cross-attention map to perform localized edits. To this end, we append a word to the prompt depending on the type of edit and inject the selected saliency-guided cross-attention map into the denoising U-Net. This process modifies \( \mathbf{Z}' \). By denoising and decoding this edited \( \mathbf{Z}'_E \), we obtain the generated image edits \( \mathbf{I}'_E \). 
We refer the reader to the supplementary material for the pseudocode of the editing algorithm.

As evidenced by cognitive studies on human visual attention\cite{Wolfe2004}, saliency is influenced by contrast, brightness, and color.  Our work involves controlled edits to these photometric properties, allowing us to control and interpret the model's saliency prediction in response to such edits.  We focus on photometric edits that enhance the saliency of the most salient region that corresponds to a word from the prompt. As a result, this region remains salient after the edit. Conversely, when an edit reduces saliency, it is unclear whether the reduction makes the region non-salient or not. We show such edits that diminish contrast and brightness and their effects on saliency in the supplementary material. We describe each edit in detail in the following section.
\subsection{Editing Types}\label{sec:editing-types}
\subsubsection{Contrast Increase.}
\begin{figure}[t]
    \centering
    \includegraphics[width=1.0\columnwidth]{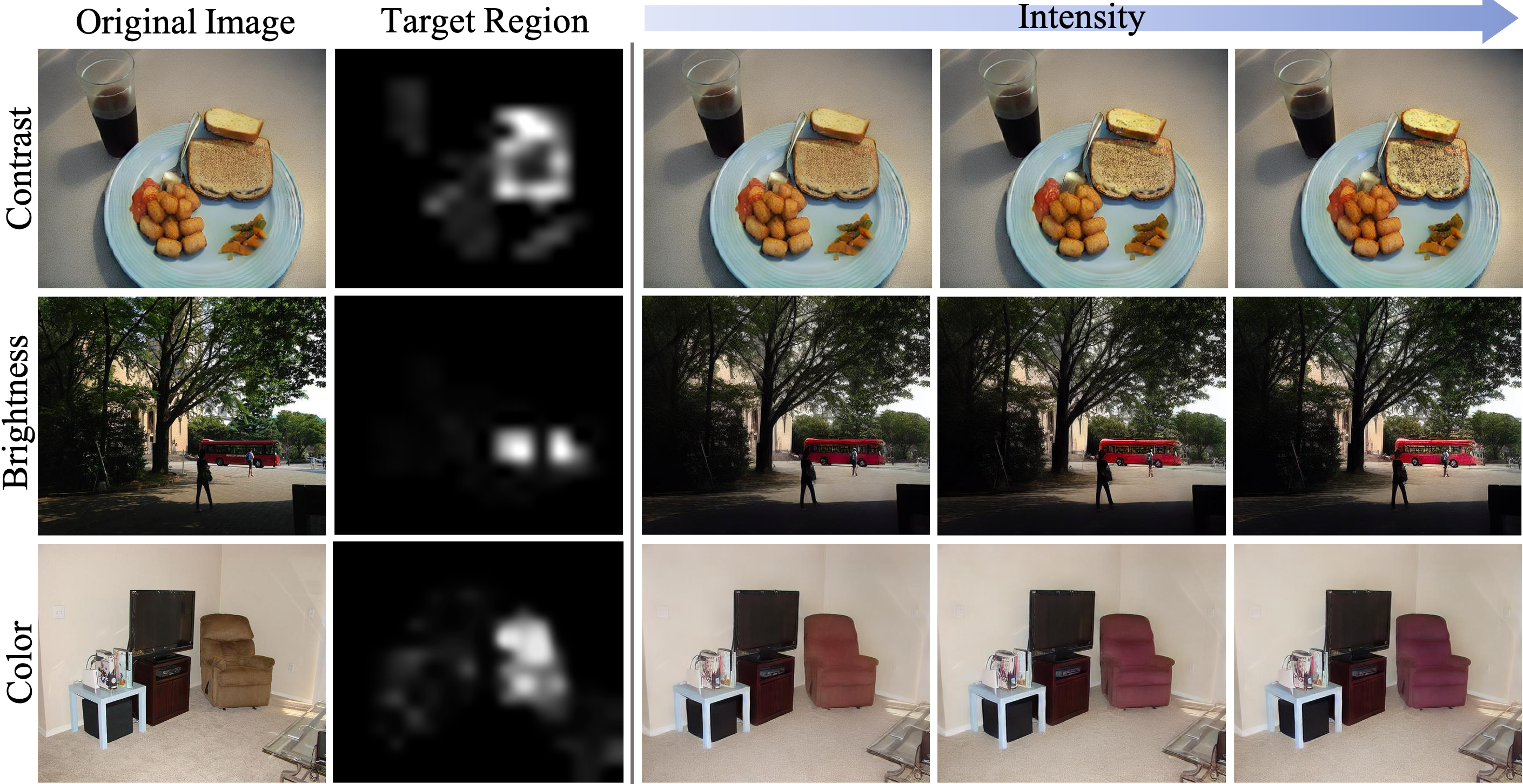}
    \caption{Original image, selected editing region, and edited images at different intensity levels for contrast, brightness, and color edits. In the first row, we increase the contrast in the bread. The second row shows an increase in the brightness of the red bus. The last row shows a progressive color edit of the chair to purple. These edits aim to increase the saliency of the target region shown in the second column. }
    \label{fig:edits-all}
\end{figure}

We increase the contrast of a specific region as 
\begin{equation}
    Z_{E} = M \times (\alpha \times (Z - \mu) + \mu) + (1-M)\times  Z\;,
\end{equation}
where $\mu$ denotes the mean inside the selected area $M$, $Z$ denotes the shared latent vector, $Z_{E}$ denotes the edited latent vector, and $\alpha$ is the scale parameter which denotes the strength of the edit. We inject the selected map into the cross-attention maps of the denoising U-Net. Following this,
we insert an empty token before the word token that corresponds to the selected region in the prompt. This approach enables altering the cross-attention while preserving the semantic content of the region. The first row in Figure \ref{fig:edits-all} shows a progressive increase in contrast at varying intensity levels.
\subsubsection{Brightness Increase.}
We increase the brightness of a specific region as
\begin{equation}
    Z_{E} = M \times (  Z + \alpha ) + (1-M)\times  Z\;,
\end{equation}
where $Z$ and $Z_{E}$ are the shared latent vector and the edited latent vector, respectively, and  $\alpha$ is the scale parameter which denotes the strength of the edit. 

Lastly, similar to the previous section, we insert an empty token into the prompt and inject the selected map into the cross-attention maps of the denoising U-Net. The second row in Figure~\ref{fig:edits-all} shows a progressive increase in brightness with varying levels of intensity.
\subsubsection{Color Change.}
To modify the color of a region, we add the target color before the word token that corresponds to the targeted region in the prompt. Then, we inject the selected map into the cross-attention maps of the denoising U-Net, thereby associating the target color with the selected regions. We scale the values of the selected map to amplify the intensity of the color change. The last row in Figure \ref{fig:edits-all} shows a gradual alteration in color at different intensity levels.
\subsection{Scaling and Constraining the Image Edits}\label{sec:constrainandmatrix}
In order to use our editing methods, we need to calculate a scale parameter because one unit increase in the RGB space does not correspond to one unit increase in the four-channel diffusion latent space. This can result in oversaturated and high-contrast images \cite{alldieck2024score}. To counter this, we introduce a \textit{channel-wise} scale parameter, $\gamma$. We start by approximating the decoder \(\mathcal{D}\) as a matrix, namely $\mathbf{A}_{4 \times 3}$, as described in~\cite{huggingface_discussion_2022}. Subsequently, we compute the pseudo-inverse of the approximated decoder matrix as $\mathbf{A^{+}}_{3 \times 4}$. We solve $\gamma_{1 \times 4}$ = $\mathbf{1}_{1 \times 3}$ $*$ $\mathbf{A^{+}}_{3 \times 4}$ to find the value of the scale parameter $\gamma_{1 \times 4}$. Consequently, we scale our image edits, specifically $\mu$ in contrast and $\alpha$ in brightness edits with this $\gamma$. We demonstrate the impact of this scaling in Section~\ref{sec:constraining}.

Additionally, we employ a mechanism to constrain edits and prevent overly strong edits. We create a vector of photometric properties for each image/patch and calculate the standard deviation across the original images. During editing, if the LLFR module's output deviates more than two standard deviations from the original image, we reduce the edit strength to maintain image naturalness. This process ensures balanced and realistic edits as we show in Section~\ref{sec:constraining}.

\subsection{Loss Functions}
We use readout, saliency, and editing losses. 
The readout losses ensure that extracted features contain the desired image properties.  We discuss the hyperparameters in the supplementary material.\\

\textbf{Readout losses}. We calculate the readout losses as the \(\mathcal{L}_{2}\) distance between the decoded image properties and their ground-truth values as:
\begin{equation}
\begin{split}
    \mathcal{L}_{\text{readout}}(I) &= \lambda_{1} * \mathcal{L}_{2}(\mu_{RGB}',\mu_{RGB})  + \lambda_{2} * \mathcal{L}_{2}(\mu_{Br}',\mu_{Br})\\ + &\mathcal{L}_{2}(c_L',c_L)  + \lambda_{3} * \mathcal{L}_{2}(c_G',c_G) +  \lambda_{4} * \mathcal{L}_{CE}(c',c)\;,
    \end{split}
\end{equation}
where $\mu_{RGB}$, $\mu_{Br}$, $c_L$, and $c_G$ are calculated as described in Section~\ref{sec:mlp-readouts}. $\mu_{Br}^{\prime}$, $c_L^{\prime}$, $c_G^{\prime}$ and $c^{\prime}$  are the predicted values shown as outputs of the LLFR and HLFR modules in Figure ~\ref{fig:train-figure} and ~\ref{fig:model-editing}.  $ \mathcal{L}_{CE}$ denotes the cross-entropy loss for classification. \\
\textbf{Saliency loss}. The saliency ground-truth maps are the blurred version of the fixation maps collected by user experiments~\cite{Judd_2009}.
Hence, they follow a spatial Gaussian distribution. The MSE denoising loss does not guarantee that the predicted saliency map follows this distribution. Hence, we use the KLD~\cite{vigkld} and CC~\cite{ccmetric} losses to improve the distribution and coverage of our prediction, respectively. Specifically, denoting the saliency ground truth for image $I$ as $S$ and the predicted saliency as $S'$, we use the loss
\begin{equation}
    \mathcal{L}_{\text{saliency}}(I) = \lambda_{5} * \mathrm{CC}(S,S') + \lambda_{6} * \mathrm{KLD}(S,S'),
\end{equation}
where 
\begin{equation}
    \text{KLD}(S',S) = \sum_{i}S_{i} \log \left(\varepsilon + \frac{S_{i}}{\varepsilon + S'_{i}}\right)\;, \label{eq:4}
\end{equation}
 with $i$ iterating over the image pixels and $\varepsilon$ being a small constant to avoid numerical instabilities. Furthermore, we have
 \begin{equation}
\text{CC} = \frac{\sum (S' - \bar{S'})(S - \bar{S})}{\sqrt{\sum (S' - \bar{S'})^2 \sum (S - \bar{S})^2}}\;,
\end{equation}
where $\bar{S}$ is the mean value over pixels and the summations run over the image pixels. CC directly measures how well the spatial distribution of the predicted saliency map matches the spatial distribution of the ground truth. \\
 \textbf{Editing Loss}. For editing, we use the loss
\begin{equation}
    \mathcal{L}_{\text{edit}}(I) = \mathrm{BCE}( \mathrm{M}\times S',\mathrm{M}\times S_{E}')\;,
\end{equation}
where BCE is the binary cross-entropy and
\begin{align}
    \mathrm{M}&= \mathrm{C}_{i^*} \times \mathbf{S}, \quad \quad i^* = \argmax_i \innerproduct{\mathrm{C}_i}{\mathbf{S}}
\end{align}
is the target editing region defined as the elementwise product of $C_{i^*}$ and $\bf{S}$ where $i^*$ is the index of the cross-attention map with the largest inner product with $\bf{S}$. $M$ represents the selected spatial attention map. We illustrate this loss in the supplementary material.
\section{Experiments and Results}
\subsection{Experimental Setup}
We train the HLFR and LLFR modules for 20k iterations to be able to reconstruct the respective image properties. We train the SR module for 22k iterations. We use a learning rate of $5\times 10^{-5}$ and $1\times 10^{-4}$ for the feature extractors and the readout networks, respectively. We use AdamW~\cite{adamw} optimizer for all modules. 
\subsection{Datasets}
We report the performance of our methods on three publicly available saliency estimation benchmarks~\cite{salicon,Judd_2009,cat2000}. We train our models on 10,000 images of the SALICON~\cite{salicon} dataset, which consists of diverse context-rich images from the MS COCO dataset~\cite{mscoco}. 
The ground truth of the official SALICON test set is not released but predictions can be submitted for evaluation on the LSUN challenge website\footnote{https://competitions.codalab.org/competitions/17136}. For the prompts, we use the captions from the MS COCO dataset~\cite{mscoco}.
\subsection{Augmenting Data}
To augment data, we use a sampling parameter $p$, which denotes the probability of training with the original image and ground truth. Otherwise, we randomly select one type of edit and train with our editing loss. We use the training images of the SALICON dataset~\cite{salicon} for augmentation. We report results with $p=0.5$ in Table~\ref{tab:sota-res} and Table~\ref{tab:sota-augs}. We compare different values of $p$ and illustrate our loss during augmentation in the supplementary material.
\begin{figure}[t]
    \centering
    \includegraphics[width=1\columnwidth]{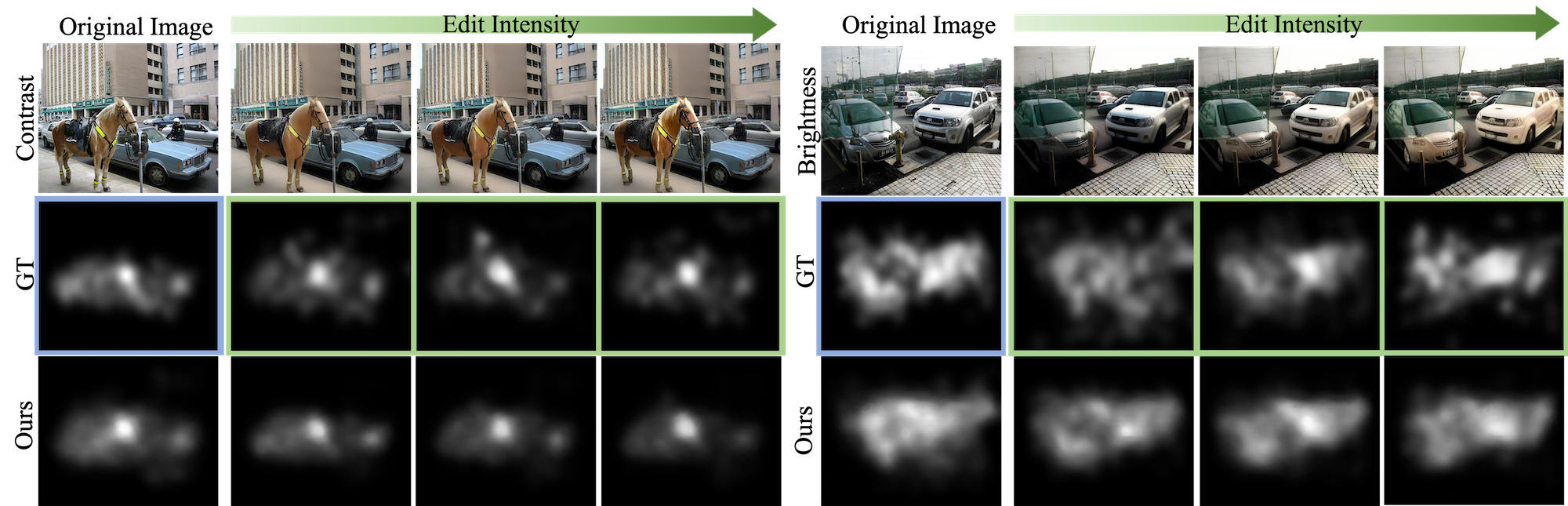}
    \caption{(\textbf{Top Row}): Original and edited images with the (\textbf{Middle Row:}) ground-truth saliency maps from SALICON~\cite{salicon}, shown in \textbf{\textcolor{mylightblue}{blue}} and from our user study, shown in \textbf{\textcolor{mygreen}{green}}. We report that our generated image edits can shift human attention toward the edited region. For instance, by enhancing the contrast of the horse in the image to the left, we observe that the attention focuses on the horse as the intensity of the edit increases. Similarly, by enhancing the brightness of the cars, they gather more attention as their brightness increases. (\textbf{Bottom Row:}) Our saliency prediction model can achieve saliency estimations that align with the ground-truth maps. }
    \label{fig:res-contrast}
\end{figure}

\subsection{User Study}
To evaluate our method, we conduct an eye-tracking user study with 8 participants. We randomly select 50 images from each editing category with three intensity levels, giving a total of 150 images. We display the images to the participants in a random sequence for 5 seconds separated by blank intervals of 2-seconds. We constructed the saliency ground truth for these generated image edits following common practice presented in~\cite{Judd_2009}. 
Figure-\ref{fig:res-contrast} presents the original and edited images alongside their saliency predictions and the collected ground truth, demonstrating our edits' effectiveness in increasing the target region's saliency. Our saliency predictions closely align with human visual attention patterns. A paired samples T-Test between the original and edited images confirms a significantly increased fixation on the edited regions, giving a p-value of $< 0.05$.

\begin{table}[t]
\centering
\scalebox{0.9}{
\setlength{\tabcolsep}{4pt}
\begin{tabular}{lccccccccc}
\multicolumn{1}{l}{ Model }  & AUC $\uparrow$ & KL $\downarrow$ & NSS $\uparrow$ & CC $\uparrow$ & SAUC $\uparrow$ & SIM $\uparrow$ \\
\cmidrule(l){1-1}\cmidrule(lr){2-7}
DINet  & 0.863 & 0.613 & 1.974 & 0.860 & 0.742 & 0.784 \\
DSCLRCN  & 0.869 & 0.637 & 1.979 & 0.831 & 0.736 & 0.715 \\
SalNet  & 0.860 & 0.674 & 1.766 & 0.730 & 0.711 & 0.696 \\
TempSAL & 0.869 & 0.195 &  1.967  & 0.911 & 0.745 & 0.800 \\
SAM  & 0.866 & 0.610 & 1.965 & 0.842 & 0.741 & 0.751 \\
SimpleNet & 0.869 &   0.201 &1.960&0.907 & 0.743 &  0.793 \\
SALICON  & 0.837 & 0.658 & 1.877 & 0.657 & 0.694 & 0.639 \\
DeepGaze IIE  & 0.869 & 0.285 & 1.996 & 0.872 & \textbf{0.767} & 0.733 \\
UNISAL   & 0.864 & 0.350 & 1.952 & 0.879 & 0.739 & 0.775 \\
RINet   & 0.869 & \textbf{0.189} & 1.982 & 0.911 & 0.746 & 0.803 \\
MDSEM   & 0.868 & 0.568 & \textbf{2.058} & 0.868 & 0.746 & 0.774 \\
\textbf{Ours} & \textbf{0.870}  & 0.191 & 1.973 & \textbf{0.914} & 0.744 & \textbf{0.805}\\

\end{tabular}
}
\caption{Evaluation results on the SALICON (LSUN 2017)
test benchmark. We compare our model with the state-of-the-art saliency prediction models, namely DINET ~\cite{dinet}, DSCLRCN ~\cite{dsclrcn}, SalNet ~\cite{salnet}, TempSAL ~\cite{aydemir2023tempsal}, SAM-ResNet~\cite{sam}, SimpleNet ~\cite{simplenet}, SALICON \cite{saliconmodel}, DeepGaze IIE ~\cite{linardos}, RINet~\cite{Song2023RINet}, MDSEM~\cite{fosco2020howmuch} and UNISAL ~\cite{unisal}. The results in bold show the best performance. Our saliency prediction method with our augmentation method outperforms the SOTA methods on 3 out of 6 metrics.}\label{tab:sota-res}
\end{table}
\subsection{Quantitative Results}
\textbf{Saliency Prediction}
We evaluate the performance of our saliency prediction model that is trained with our data augmentation method and tested on the SALICON test data~\cite{salicon}. Table~\ref{tab:sota-res} compares standard evaluation metrics for different state-of-the-art saliency models alongside our model. Our model outperforms the SOTA models on 3 out of 6 metrics. We present additional results on MIT1003 and CAT2000 datasets in the supplementary material.\\
\textbf{Data Augmentation}
We evaluate the performance of our data augmentation method by choosing the best-performing baseline models and training on the SALICON~\cite{salicon} dataset. 
We report the saliency prediction performance on the SALICON validation set\cite{salicon}. 
Table~\ref{tab:sota-augs} compares standard evaluation metrics for different baseline saliency models alongside our method with and without our data augmentation technique. Our augmentation method consistently improves the saliency prediction performance of all the baseline models. We present additional results on MIT1003 and CAT2000 datasets in the supplementary material.
\begin{table}[t]
\centering
\centering
\setlength{\tabcolsep}{5pt}
\scalebox{0.9}{
\begin{tabular}{lccccccccccccccc}
\multicolumn{1}{l}{  } & \multicolumn{4}{c}{ w/o Augmentation }  & \multicolumn{4}{c}{ w/ Augmentation } \\
\multicolumn{1}{l}{ Model }   & KL $\downarrow$ & NSS $\uparrow$ & CC $\uparrow$ &SIM $\uparrow$ & KL $\downarrow$ & NSS $\uparrow$ & CC $\uparrow$  & SIM $\uparrow$ \\
\cmidrule(l){1-1}\cmidrule(lr){2-5}\cmidrule(lr){6-9}

SimpleNet  &   0.193 &1.926&0.907 &  0.797 &   0.185 &1.937 &0.911 &  0.805 \\
TempSAL  & 0.198 &  1.930  & 0.906 &\textbf{ 0.798} &   0.181 &1.949 &0.911 &  0.802 \\
DeepGaze IIE  & 0.314 & \textbf{1.954} & 0.872 & 0.733  & 0.258 & \textbf{1.972} & 0.883 & 0.749 \\
UNISAL   & 0.226 & 1.923 & 0.880  & 0.771 & 0.212 & 1.930 & 0.894  & 0.797\\

\textbf{Ours}  & \textbf{0.191} & 1.927 & \textbf{0.908}  & 0.788 & \textbf{0.179} & 1.946 & \textbf{0.915}  & \textbf{0.807}\\
\end{tabular}}
\caption{Evaluation results on the SALICON validation set~\cite{salicon}. We select four existing saliency prediction models as baselines, namely SimpleNet ~\cite{simplenet}, DeepGaze IIE ~\cite{linardos}, TempSAL ~\cite{aydemir2023tempsal} and UNISAL ~\cite{unisal}. We train all models with the SALICON~\cite{salicon} dataset with and without our augmentation method. Our augmentation method consistently improves the saliency prediction performance of all models, in all metrics. Additionally, our saliency prediction method with data augmentation outperforms the other methods in 3 out of 4 metrics. The results in bold show the best performance.}\label{tab:sota-augs}
\end{table}
\subsection{Ablation Studies} 
\textbf{Comparison with Classical Augmentation Methods}
We evaluate our augmentation method against 12 classical augmentation methods namely, rotation, vertical flip, horizontal flip, cropping, JPEG compression, motion blur, inversion, noise, contrast, and shearing. We apply the same transformation to the image and its ground truth saliency map. We train all models on the SALICON dataset ~\cite{salicon} and the augmented images with these methods. We observe that most of the augmentations decrease performance with the exception of horizontal flip. We provide a table of all parameters and results in the supplementary material.\\
\begin{figure}[tb]
    \centering
    \includegraphics[width=0.9\columnwidth]{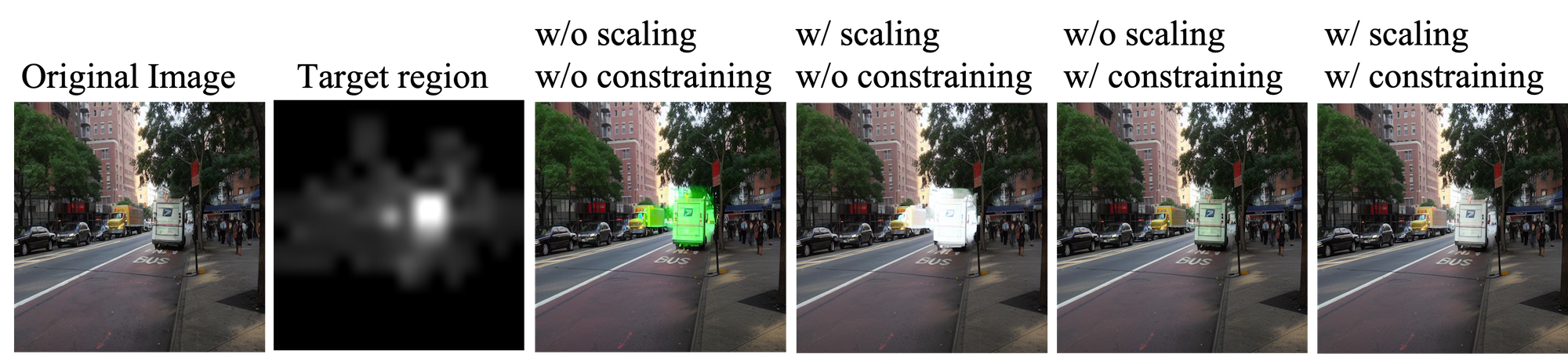} 
    \caption{Scaling ensures that edits within the latent space are equivalent to those in RGB space and constraining the edits prevents the generation of excessively strong edits. For instance, the green coloration in the edited regions indicates the absence of scaling, and without proper constraints, edits can become excessively strong, leading to unnatural results.
     }
    \label{fig:ablation-green}
\end{figure}
\textbf{Effect of Scaling and Constraining the Image Edits}\label{sec:constraining} We show qualitative ablations of the effect of scaling and constraining the image edits in Figure~\ref{fig:ablation-green}. In the absence of scaling, and without proper constraints, edits can become excessively strong, leading to unnatural results. \\
We provide additional qualitative and quantitative results, ablation on the augmentation methods, and losses in the supplementary material.

\section{Conclusion}
We have introduced a novel data augmentation approach for deep saliency prediction, addressing the challenge of limited labeled data diversity and quantity. We perform targeted edits on photometric properties such as contrast, brightness, and color while preserving the complexity and variability of real-world visual scenes. Our experiments have concluded that our data augmentation method is suitable for saliency prediction. 
 Moreover, leveraging these features that generate augmentation for saliency prediction yields a better understanding of visual attention patterns as shown by a user study. We hope that this work will contribute to advancing the field of saliency prediction through our data augmentation method, characterized by its ability to generate highly relevant and diverse training examples.  

\section*{Acknowledgement}
This work was supported in part by the Swiss National Science Foundation via the Sinergia grant CRSII5-180359.

\bibliographystyle{splncs04}
\bibliography{main}
\end{document}